\documentclass[letterpaper]{article} 
\usepackage[]{aaai2026}  
\usepackage{times}  
\usepackage{helvet}  
\usepackage{courier}  
\usepackage[hyphens]{url}  
\usepackage{graphicx} 
\usepackage{natbib}  
\usepackage{caption} 
\usepackage{algorithm}
\usepackage{algorithmic}

\usepackage{amssymb}
\usepackage{amsmath} 

\usepackage{multirow}
\usepackage{graphicx}
\usepackage{booktabs}
\usepackage{tabularx}

\usepackage[table]{xcolor}

\usepackage{newfloat}
\usepackage{listings}
\DeclareCaptionStyle{ruled}{labelfont=normalfont,labelsep=colon,strut=off} 
\floatstyle{ruled}
\newfloat{listing}{tb}{lst}{}
\floatname{listing}{Listing}
\title{WinMamba: Multi-Scale Shifted Windows in State Space Model\\for 3D Object Detection}

\author{
    Longhui Zheng\textsuperscript{\rm 1,\rm 2}\equalcontrib, 
    Qiming Xia\textsuperscript{\rm 1,\rm 2}\equalcontrib,
    Xiaolu Chen\textsuperscript{\rm 1,\rm 2}, 
    Zhaoliang Liu\textsuperscript{\rm 1,\rm 2},
    Chenglu Wen\textsuperscript{\rm 1,\rm 2}\textsuperscript{}\thanks{Corresponding authors.}
}

\affiliations{
    \textsuperscript{\rm 1}Fujian Key Laboratory of Sensing and Computing for Smart Cities, Xiamen University, China \\
    \textsuperscript{\rm 2}Key Laboratory of Multimedia Trusted Perception and Efficient Computing, \\Ministry of Education of China, Xiamen University, China  \\
}

\usepackage{bibentry}

\begin{document}

\maketitle

\begin{abstract}
3D object detection is critical for autonomous driving, yet it remains fundamentally challenging to simultaneously maximize computational efficiency and capture long-range spatial dependencies.
We observed that Mamba-based models, with their linear state-space design, capture long-range dependencies at lower cost, offering a promising balance between efficiency and accuracy.
However, existing methods rely on axis-aligned scanning within a fixed window, inevitably discarding spatial information. 
To address this problem, we propose WinMamba, a novel Mamba-based 3D feature-encoding backbone composed of stacked WinMamba blocks. 
To enhance the backbone with robust multi-scale representation, the WinMamba block incorporates a window-scale-adaptive module that compensates voxel features across varying resolutions during sampling. 
Meanwhile, to obtain rich contextual cues within the linear state space, we equip the WinMamba layer with a learnable positional encoding and a window-shift strategy.
Extensive experiments on the KITTI and Waymo datasets demonstrate that WinMamba significantly outperforms the baseline. Ablation studies further validate the individual contributions of the WSF and AWF modules in improving detection accuracy. The code will be made publicly available.
\end{abstract}

\section{Introduction}

3D object detection, serving as a core task for 3D scene perception, finds extensive applications in domains such as robotic navigation and autonomous driving. However, scene perception faces numerous challenges, among which the critical trade-off between computational complexity and long-range dependency modeling stands out as particularly severe in 3D contexts.

Recently, feature extractors based on the Mamba architecture have achieved remarkable progress in NLP and 2D computer vision. 
Owing to its powerful modeling capabilities and linear computational complexity, Mamba promises to be an effective solution to the aforementioned problems. 
Consequently, some researchers have dedicated efforts to adapting the Mamba model to the 3D domain. 
Specifically, Voxel mamba\cite{zhang2024voxel} proposes a group-free strategy, serializing the entire voxel space into a single sequence to minimize the loss of spatial proximity information. 
In contrast, UniMamba\cite{jin2025unimamba} introduces a unified Mamba architecture that efficiently achieves the simultaneous aggregation of both local and global spatial contexts. 
They both have introduced state space models into the 3D domain and achieved promising results.

\begin{figure}[t]
\centering
\includegraphics[width=1\columnwidth]{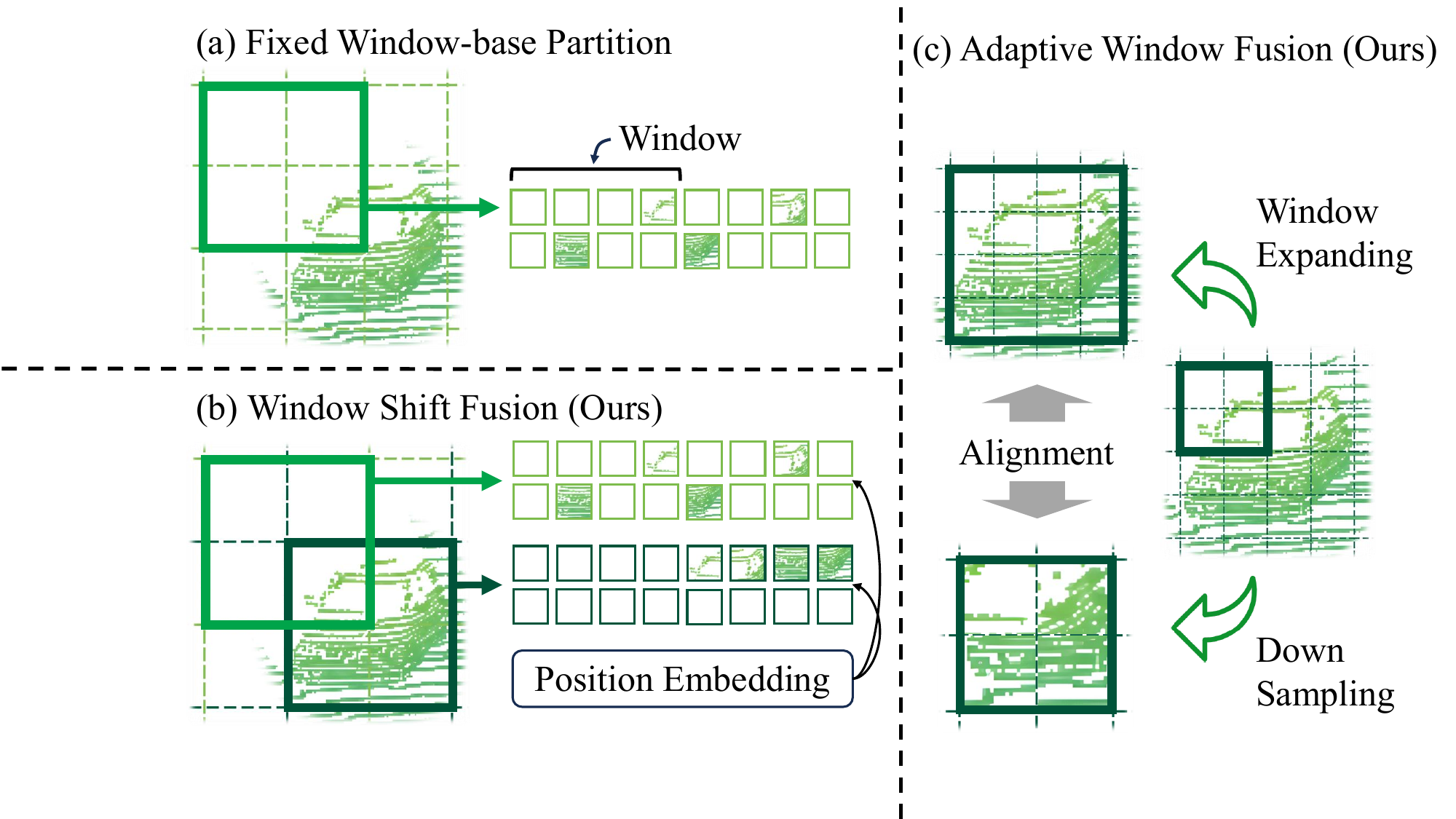}
\caption{
Motivation \& Solution
(a) causes spatial fragmentation during serialization due to fixed-window partitioning.
(b) recovers cross-window contextual information by WSF.
(c) aligns window scales with feature pyramid and preserves semantic coherence within windows through AWF.
}
\label{fig1}
\end{figure}

Although the aforementioned methods have achieved certain success in the field of 3D object detection, they also reveal several critical issues. 
When applying linear models to multi-dimensional data, serialization is often necessary. For example, 2D images may be arranged into pixel sequences, while 3D point clouds are transformed into voxel sequences.
However, this serialization process often leads to the loss or distortion of the original spatial structural information. 
Although traditional window-based serialization (first partitions the space into windows and then scans along fixed axes) does capture local features and thus mitigates the issue, it inevitably disrupts spatial continuity.

As shown in Figure~\ref{fig1}(a), the objects being segmented and dispersed across two or even multiple windows. 
After serialization, information pertaining to these elements becomes fragmented, and spatial structural information is consequently distorted.
Meanwhile, we observe that mainstream backbones typically adopt a Feature-Pyramid Networks (FPN), where multi-scale features are extracted through upsampling and downsampling operations. 
However, repeated downsampling nearly obliterates the fine structures of target objects. 
Fixed-size windows confine the model to coarse feature representations, severely impairing precise localization and recognition of detail-sensitive targets.
These two problems prompt us to consider: \textit{How can we preserve the continuity of features in feature space after they have been serialized? 
And how can we maintain alignment among features within corresponding windows during multi-scale feature encoding?}

In this paper, we propose a novel Mamba-based 3D feature-encoding backbone composed of stacked WinMamba blocks.
To address weak cross-window correlations, we propose Window Shift Fusion (WSF). As shown in Figure~\ref{fig1}(b), this method adjusts serialization outcomes by shifting window partitions, then fuses features before and after shifting. WSF enables denser window coverage across the entire space, effectively alleviating object discontinuity caused by rigid window partitioning.
Furthermore, to reduce detail degradation in feature aggregation, we introduce Adaptive Window Fusion (AWF). 
It enhances multi-scale representation by constructing an auxiliary path with higher-resolution features.
As shown in Figure~\ref{fig1}(c), AWF aligns semantic units across windows, by dynamically adjusting window sizes based on feature scales.
The fused features compensate for information loss from single-scale processing, synergistically boosting detection accuracy. 

To validate the efficacy of WinMamba, we conduct extensive experiments on the KITTI and Waymo benchmarks, complemented by ablation studies. Below, we summarize our key contributions:

\begin{itemize}
\item We propose WinMamba. A simple yet effective window-based 3D feature-encoder backbone built on Mamba architecture, designed to enhance linear models' capability for multidimensional spatial modeling.
\item We introduce Window Shift Fusion (WSF). It compensates for the cross-window context loss caused by rigid partitioning through window shifting and feature fusion.
\item We present Adaptive Window Fusion (AWF). It constructs a higher-resolution auxiliary path with adaptive window sizing to align semantic units, fusing features to recover lost details.
\end{itemize}

\section{Related Work}

\subsection{3D Object Detection in Point Clouds.}
The mainstream approaches can be broadly categorized into point-based and voxel-based methods. 
Point-based methods \cite{chen2019fast, yang2019std, qi2019deep, cheng2021back,liu2021group, pan20213d, he2020structure, shi2019pointrcnn, zhang2022not, yang20203dssd, qi2018frustum, yang2022dbq, chen2022sasa} typically operate directly on raw point clouds, employing sampling techniques and point encoders \cite{qi2017pointnet, qi2017pointnet++} to extract per-point features. 
However, the core operations of point sampling and grouping in these methods are often computationally intensive, limiting their efficiency. 
In contrast, voxel-based methods \cite{dong2022mssvt, deng2021voxel, liu2020tanet, shi2020pv, shi2023pv, shi2020points, guan2022m3detr, wang2022cagroup3d, yan2018second, yin2021center, yang2023pvt, zhang2024safdnet, zhao2025sp3d, xia20233, ye2025fade3d} discretize the unstructured point clouds into regular 3D voxel grids, leveraging 3D sparse convolutions for efficient spatial feature extraction. 
Although voxel-based approaches demonstrate superior performance, their inherent limitation stems from the local receptive fields of 3D convolutions, which may struggle to adequately model long-range dependencies. 
To overcome this limitation, recent studies\cite{chen2023largekernel3d, lu2023link} have explored strategies employing large-kernel convolutions, aiming to significantly expand the receptive field and thereby enhance detection accuracy.

\subsection{Mamba in the Vision Task.}
Mamba\cite{gu2023mamba}, a state space model (SSM) with linear complexity, has emerged as an efficient alternative to Transformers for sequence modeling. Its application in vision tasks is gaining momentum. 
Recently, The extension of Mamba to vision tasks has been explored in several works\cite{huang2024localmamba, liu2024vmamba}. 
In 2D vision, methods like Vim\cite{zhu2024vision} and GrootVL\cite{xiao2024grootvl} explore different scanning strategies, such as bidirectional SSMs and tree-based serialization, to adapt Mamba for image feature extraction. 
Recently, Mamba has also been introduced to 3D object detection. 
Voxel Mamba\cite{zhang2024voxel} employs a group-free voxel serialization strategy to preserve spatial proximity, and introduces a Dual-scale SSM Block for hierarchical context modeling.
UniMamba\cite{jin2025unimamba} further integrates 3D convolution with SSM using a unified block. It combines Z-order serialization and local-global aggregation to efficiently capture spatial-channel dependencies. 
These works demonstrate Mamba’s potential as a powerful and scalable alternative to Transformers in both 2D and 3D vision tasks.

\section{Method}
\begin{figure*}[t]
\centering
\includegraphics[width=1\textwidth]{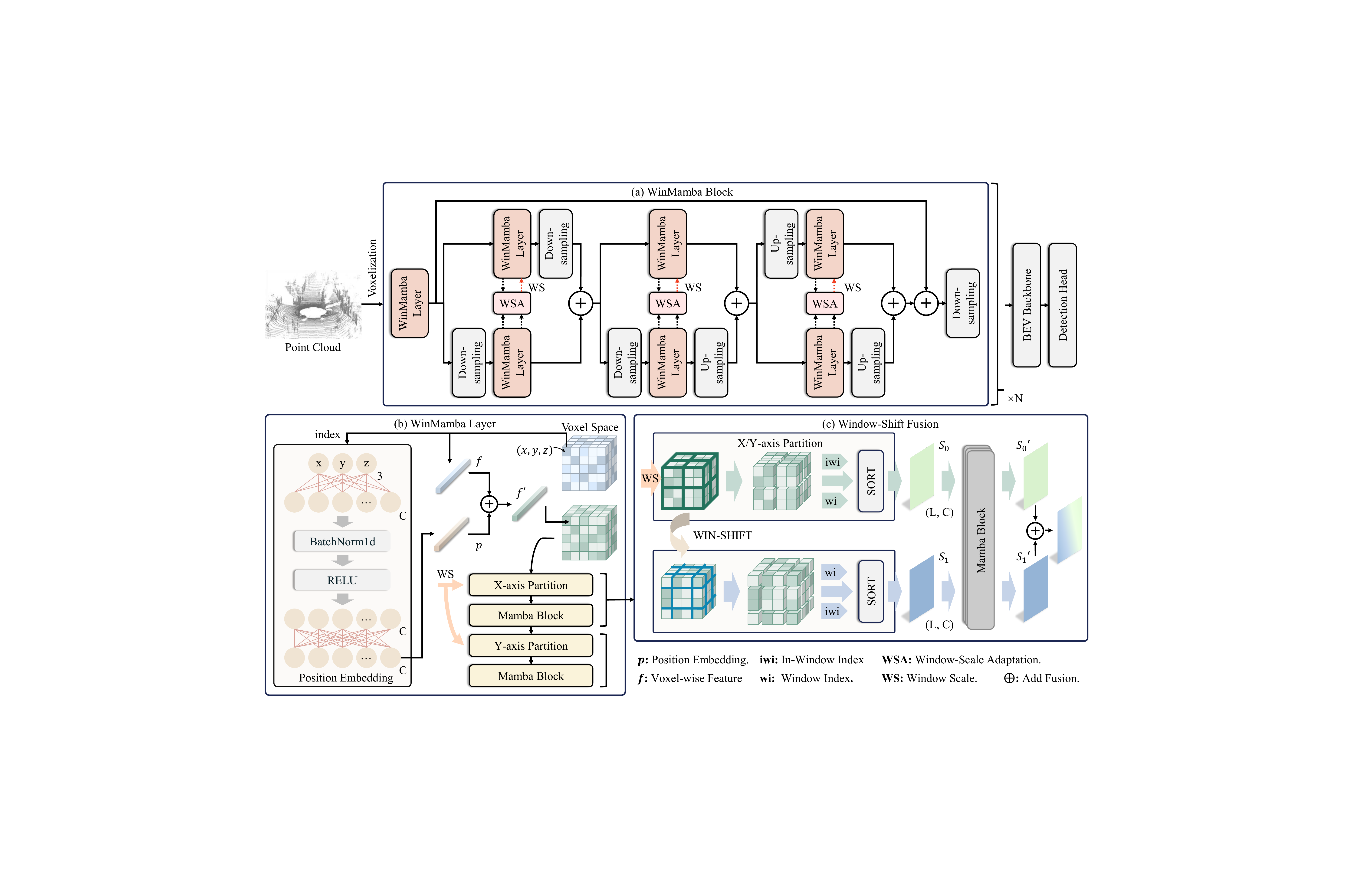}
\caption{
(a) illustrates the structure of WinMamba, composed of N stacked WinMamba Blocks. The lower path functions as the main path (FPN), while the upper path serves as the auxiliary path. Between the two paths, Window-Scale Adaptation (WSA) dynamically assigns suitable window sizes to the auxiliary path.
(b) details the WinMamba layer, a feature extraction module incorporating learnable positional embeddings. The input features undergo alternating window partitioning along the X- and Y-axis, then be processed by the Mamba Block.
(c) presents the Window Shift Fusion process. Voxel-wise features are serialized from both pre- and post-shift window partitions, encoded by the Mamba block, and subsequently fused.
}
\label{fig2}
\end{figure*}

This paper introduces WinMamba, a novel 3D feature-encoder backbone based on Mamba. As illustrated in Figure~\ref{fig2}, WinMamba comprises two key modules:
(1) Adaptive Window Fusion (AWF);
(2) Window Shift Fusion (WSF).
We first present an overview of the WinMamba architecture, and then provide detailed descriptions of the proposed window-based modules.

\subsection{Mamba-Based 3D Feature-Encoding Backbone}

\textbf{Overview of WinMamba.} 
WinMamba is intended to efficiently and adaptively extract multi-scale features, preserving complete spatial structural information. 
The 3D backbone is built upon the Mamba architecture, incorporating two well-designed window-based feature enhancement strategies.
As shown in Figure~\ref{fig2}(a), WinMamba follows the standard paradigm of mainstream voxel-based detectors\cite{wu2022casa}.
It comprises four modules: Voxel Feature Encoder, 3D Backbone Network, BEV Backbone Network, and Detection Head. 
Our core innovation lies in the 3D backbone network, constructed by multiple stacked WinMamba blocks. The details are elaborated in the following sections.

\textbf{WinMamba Layer.} 
The WinMamba Layer serves as the crucial component of the WinMamba Block, integrating a Position Embedding Layer, Window Partitions, and Mamba Blocks. 
As shown in Figure~\ref{fig2}(b), the process begins by feeding the index $e\in\mathbb{R}^{3}$ of the voxel-wise features $f\in\mathbb{R}^{C}$ into the position embedding layer. 
The embeddings are fused element-wise with the original $f$ to produce enhanced voxel-wise features $f'$.
Then, the features undergo alternating axial feature interactions: sequentially processed through X-axis partition, followed by the Mamba Block, then Y-axis partition, followed by another Mamba Block.
This iterative multi-axial mechanism comprehensively explores spatial contextual information across orthogonal dimensions.

\textbf{Position Embedding Layer.} 
To encode the spatial information of the voxel-wise feature, we propose a learnable position embedding layer. 
This layer maps the index $e$ of each voxel into a position embedding $p\in\mathbb{R}^{C}$ that shares the same dimensionality as the voxel-wise features $f$. 
Specifically, as illustrated in Figure~\ref{fig2}(b), we employ a multi-layer perceptron consisting of two linear transformations to perform this nonlinear mapping:
\begin{equation}
    \begin{aligned}
        p&=MLP(e)\\
        &=\omega_{2}(ReLu(BN(\omega_{1}e+b_{1})))+b_{2},
    \end{aligned}
\end{equation}
where $\omega_{1}$ and $\omega_{2}$ are learnable weight matrices,
$b_{1}$ and $b_{2}$ are learnable bias matrices,
$BN$ denotes batch normalization, 
and $ReLU$ is the activation function. 

Then, we integrate the position embedding with the corresponding voxel-wise feature via element-wise addition to incorporate spatial priors. The resulting feature $f'$ retains the original semantic information while explicitly encoding spatial positional relationships, thus providing richer contextual representations for subsequent network processing.

\subsection{Adaptive Window Fusion (AWF)}

\begin{figure}[t]
\centering
\includegraphics[width=1\columnwidth]{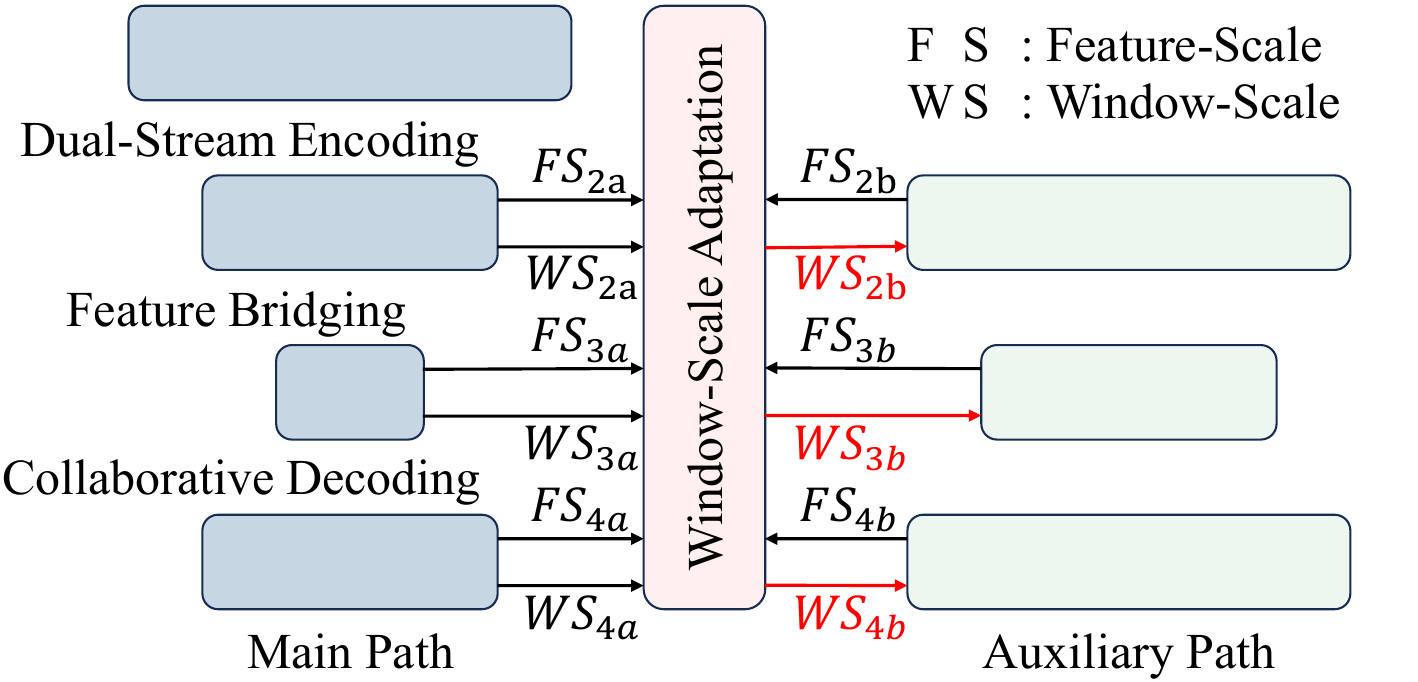}
\caption{Illustration of the scale alignment process between Main Path and Auxiliary Path across different feature levels.}
\label{fig3}
\end{figure}

To address the issue of detail loss during the feature aggregation based on downsampling in traditional FPN, we propose a novel method called Adaptive Window Fusion.
AWF is designed around the concept of multi-path parallel architecture. 
It explicitly associates different FPN levels with corresponding window scales, allowing the model to learn highly consistent semantic information across scales.

\textbf{Window-Scale Adaptation (WSA)}
To ensure semantic alignment of features within their respective windows between corresponding feature levels of the main and auxiliary paths, we propose Window-Scale Adaptation. 
This module takes the feature scales (FS) of both paths and the window scale (WS) of the main path as input, then computes the required window scale for the auxiliary path according to the following principle:
\begin{equation}
    \frac{FS_{a} }{FS_{b} } =  \frac{WS_{b} }{WS_{a} },
\end{equation}
where the subscript $a$ denotes the main path and $b$ denotes the auxiliary path.

\textbf{WinMamba Block.}
The WinMamba Block is built upon the main path of a standard FPN structure. 
In parallel, it incorporates an auxiliary path whose feature level is higher than the main path. 
The main path employs a fixed window size $ws_{a}$, while the window size $ws_{b}$ of the auxiliary path is dynamically set via WSA. 
As shown in Figure~\ref{fig3}, the WinMamba Block comprises three core components: Dual-stream Encoding, Feature Bridging, and Collaborative Decoding. 
Following these components, the module reduces feature dimensionality through a downsampling layer. 
This processed output is then passed to subsequent blocks, enabling the progressive construction of BEV features through multiple stacked WinMamba Blocks.

\textbf{Dual-Stream Encoding.}
In the dual-stream encoding structure, the main path first performs a downsampling operation, followed by feature extraction using a WinMamba Layer. 
Feature interaction occurs after downsampling, where the spatial resolution is reduced to $1⁄d$ of the input.
In contrast, the auxiliary path applies the WinMamba Layer prior to downsampling, enabling feature interaction to occur at the original spatial resolution. 
To ensure semantic alignment with the corresponding window in the main path, we apply WSA, which adaptively enlarges the window size in the auxiliary path to $d \cdot ws$.
After feature extraction in both paths, outputs are finally aggregated via a fusion operation.
\begin{equation}
\left\{\begin{matrix}
\begin{aligned}
    O_{1a} &= WL(D(I_a, d), ws),\\
    O_{2a} &= D(WL(I_a, d \cdot ws), d),\\
    O_a &= O_{1a} \oplus O_{2a},
\end{aligned}
\end{matrix}\right.
\end{equation}
where $I_a$ denotes the input feature, 
$O_a$ represents the output feature, 
$D(\cdot, d)$ is the downsampling operation with factor d, $WL(\cdot, ws)$ denotes the WinMamba Layer with window size $ws$, 
and $\oplus$ indicates element-wise addition.

\textbf{Feature Bridging.} 
In the feature bridging structure, the main path first performs a downsampling operation, followed by a WinMamba Layer, and then applies an upsampling operation. 
Feature interaction occurs after downsampling, where the spatial resolution is reduced to $1/d$ of the input. 
In contrast, the auxiliary path only involves a WinMamba Layer, with feature interaction occurring before downsampling. 
As a result, the features in this path retain their original spatial resolution during interaction. 
To ensure semantic alignment with the corresponding window in the main path, we apply WSA, which adaptively enlarges the window size in the auxiliary path to $d \cdot ws$.
After feature extraction in both paths, their outputs are finally aggregated via a fusion operation
\begin{equation}
\left\{\begin{matrix}
\begin{aligned}
    O_{1b} &= U(WL(D(I_b,d),ws),u),\\
    O_{2b} &= WL(I_b,d\cdot ws),\\
    O_b &= O_{1b}\oplus O_{2b},
\end{aligned}
\end{matrix}\right.
\end{equation}
where $I_b$ denotes the input feature, 
$O_b$ is the output feature, 
and $U(\cdot, u)$ is the upsampling operation with factor $u$.

Since the auxiliary path does not involve any spatial scaling operations at this part, it effectively preserves most of the spatial details from the input feature $I_b$.
This allows the auxiliary output $O_{2b}$ to naturally serve a role similar to residual connections in traditional FPNs, where original spatial information is propagated. 
According to this, we directly replace the explicit residual connection with the auxiliary path. 
This replacement not only simplifies the overall structure, but also avoids potential information redundancy arising from additional residual connection. 

\textbf{Collaborative Decoding.}
In the collaborative decoding structure, the main path first processes the input using a WinMamba Layer, followed by an upsampling operation. 
Feature interaction occurs before upsampling, meaning the spatial resolution remains unchanged during interaction. 
In contrast, the auxiliary path first applies an upsampling operation, followed by a WinMamba Layer.
In this case, feature interaction takes place after upsampling, and the spatial resolution is therefore enlarged by $u$. 
To ensure semantic alignment within the corresponding window in the main path, we apply WSA, which adaptively enlarges the window size in the auxiliary path to $u \cdot ws$.
After feature extraction in both paths, outputs are finally aggregated via a fusion operation
\begin{equation}
\left\{\begin{matrix}
\begin{aligned}
    O_{1c} &= U(WL(I_c,ws),u),\\
    O_{2c} &= WL(U(I_c,u),u \cdot ws),\\
    O_c &= O_{1c} \oplus O_{2c},
\end{aligned}
\end{matrix}\right.
\end{equation}
where $I_c$ denotes the input feature, 
and $O_c$ is the output.

AWF employs a parallel architecture design, building associations between window scales and feature levels. 
This allows windows of different sizes to precisely align with the feature resolutions at each FPN level, thereby ensuring highly consistent semantic information within corresponding windows across paths. 
This mechanism effectively mitigates the information attenuation caused by single-scale window modeling in cross-scale feature extraction, and ultimately achieve collaborative optimization of cross-scale feature representation.

\subsection{Window Shift Fusion (WSF)}

Although traditional fixed window partitioning enhances computational efficiency, the rigid segmentation boundaries significantly weakens the spatial correlations between voxels in adjacent windows. 
This is particularly detrimental for objects spanning window boundaries, as their geometric integrity becomes vulnerable to fragmentation. 
To compensate for spatial information loss, we propose Window Shift Fusion. It shifts windows to create interleaved, complementary coverage, resulting in denser coverage across the spatial.

To construct interleaved windows for compensation, we introduce preset offset vectors $(\Delta_{x}, \Delta_{y}, \Delta_{z})$ along the spatial axes.
The original voxel coordinates $(x_{0}, y_{0}, z_{0})$ are transformed as:
\begin{equation}
    (x_{1}, y_{1}, z_{1}) = (x_{0} + \Delta_{x}, y_{0} + \Delta_{y}, z_{0} + \Delta_{z}),
\end{equation}

As a linear 3D detector, WinMamba begins with voxel serialization. Based on 3D Sparse Window Partitioning, the process consists of three steps:
1) Voxelization: Convert input point clouds into voxels.
2) Window Partitioning: Divide voxels into 3D windows of shape $(w_{x}, w_{y}, w_{z})$. 
3) Voxel Sorting: Sort voxels along the X/Y-axis based on their positional information, including the window index $wi$ and the in-window index $iwi$.:
\begin{equation}
    S, idx = Sort(V, k),\\
    \left\{\begin{matrix}
    \begin{aligned}
        W &= w_{x} \cdot w_{y} \cdot w_{z},\\
        k &= wi \cdot W + iwi,
    \end{aligned}
    \end{matrix}\right.
\end{equation}
where $W$ denotes the window size, and $Sort(\cdot, k)$ denotes sorting the input based on the key $k$. To facilitate the inverse mapping from sequence back to voxel space, we retain the indices $idx$. 
Then, we compute the sequences for both the original($S_{0}$) and shifted($S_{1}$) windows.
The concatenated sequence $S_{t}$ is fed into a Mamba Block for feature extraction:
\begin{equation}
\begin{aligned}
    S_{t} &= Concat((S_{0}, S_{1}), seq)),\\
    S_{t}' &= Mamba(S_{t}),
\end{aligned}
\end{equation}
where $Concat(\cdot, dim)$ denotes concatenation along dimension $dim$, $seq$ indicates the sequence dimension, and $Mamba$ refers to the Mamba Block.
The sequence $S_{t}'$ then split along the sequence dimension to obtain the original feature $S_{0}'$ and the shifted feature $S_{1}'$. 
\begin{equation}
    [S_{0}', S_{1}'] = S_{t}',
\end{equation}
where $\left[\cdot, \cdot\right]$ denotes the binary split operation along the sequence dimension.

Following that, the two sequences are individually mapped back to their original positions in the voxel space using the recorded index $idx$.
The final output feature $F_{t}$ is then obtained by performing element-wise summation of the two reconstructed voxel features, effectively fusing the information from both original and shifted windows.
\begin{equation}
\begin{aligned}
    &F = Map(S, idx),\\
    &F_{t} = F_{0} \oplus F_{1},
\end{aligned}
\end{equation}
where $F$ denotes the feature representation in voxel space, 
$Map(\cdot,idx)$ indicates the mapping operation using index $idx$,
$\oplus$ represents element-wise summation.

In summary, WSF mitigates the spatial discontinuity caused by rigid window partitioning in voxelized 3D representations. 
By introducing shifted windows interleaved with the original ones, WSF enables complementary spatial coverage, effectively bridging features across window boundaries. 
This design allows the model to better capture the geometric continuity of objects, especially those spanning multiple windows, while preserving the computational advantages of sparse window partitioning. 
Overall, WSF serves as a simple yet effective strategy to enhance cross-window feature interaction without compromising efficiency.

\section{ Experiments}

\subsection{Datasets and Metrics}

\begin{table*}[t]
\centering
\setlength{\tabcolsep}{1.9mm}
\begin{tabularx}{\textwidth}{l|c|ccc|ccc|ccc|c}
\toprule
\multirow{2}{*}{Method} & \multirow{2}{*}{Present at} & \multicolumn{3}{c|}{Car} & \multicolumn{3}{c|}{Pedestrian} & \multicolumn{3}{c|}{Cyclist} & \multirow{2}{*}{mAP} \\
 &  & Easy & Moderate & Hard & Easy & Moderate & Hard & Easy & Moderate & Hard &  \\ \midrule
VoxelNet & CVPR 18 & 77.5 & 65.1 & 57.7 & 39.5 & 33.7 & 31.5 & 61.2 & 48.4 & 44.4 & 51.0 \\
SECOND & Sensors 18 & 83.1 & 73.7 & 66.2 & 51.1 & 42.6 & 37.3 & 70.5 & 53.9 & 46.9 & 58.4 \\
PointPillars & CVPR 19 & 79.1 & 75.0 & 68.3 & 52.1 & 43.5 & 41.5 & 75.8 & 59.1 & 52.9 & 60.8 \\
PointRCNN & CVPR 19 & 85.9 & 75.8 & 68.3 & 49.4 & 41.8 & 38.6 & 73.9 & 59.6 & 53.6 & 60.8 \\
TANet & AAAI 20 & 83.8 & 75.4 & 67.7 & 54.9 & 46.7 & 42.4 & 73.8 & 59.9 & 53.5 & 62.0 \\
PillarNet & ECCV 22 & 87.3 & 77.7 & 76.6 & 47.9 & 43.8 & 41.3 & 80.0 & 60.9 & 57.9 & 63.7 \\
DSVT-Pillar & CVPR 23 & 87.3 & 77.4 & 76.2 & 61.4 & 56.8 & 51.8 & 82.3 & 67.1 & 63.7 & 69.3 \\
DSVT-Voxel & CVPR 23 & 87.8 & 77.8 & 76.8 & 66.1 & 59.7 & 55.2 & 83.5 & 66.7 & 63.2 & 70.8 \\ \midrule
LION-Mamba & NeurIPS 24 & 88.1 & 77.8 & 76.7 & 64.8 & 59.3 & 54.9 & 83.9 & 67.3 & 64.1 & 70.8 \\
WinMamba(Ours) & - - & \textbf{88.4} & \textbf{78.2} & \textbf{77.1} & \textbf{69.3} & \textbf{63.3} & \textbf{57.7} & \textbf{90.1} & \textbf{71.4} & \textbf{67.3} & \textbf{73.7} \\ \bottomrule
\end{tabularx}
\caption{Comparison of model performance on the KITTI validation set. all difficulties are calculated with recall 11}
\label{table1}
\end{table*}

\textbf{KITTI Dataset.} 
Widely recognized as a benchmark in autonomous driving research, the KITTI dataset captures real-world road scenes via synchronized multimodal data acquisition, using an onboard stereo camera and a 64-beam LiDAR system.
The 3D object detection subset comprises $ \sim $7.5K annotated training frames and a corresponding test set, with precise 3D bounding box annotations for three target classes: Car, Pedestrian, and Cyclist.
Each object category is evaluated using Average Precision (AP), following the KITTI benchmark’s standard metrics.
AP is defined as the area under the Precision-Recall curve, computed in either BEV or 3D space with class-specific IoU thresholds (Car: 0.7, Pedestrian / Cyclist: 0.5).
Results are reported according to the official difficulty levels (Easy, Moderate, and Hard), which are determined based on object size, occlusion, and truncation. 
We evaluate our method on the KITTI dataset using mAP computed over all classes and all difficulty levels. The AP is calculated following R11.

\textbf{Waymo Open Dataset (WOD).} 
The Waymo Open Dataset represents a more recent benchmark, featuring significantly larger scale and greater diversity of driving scenarios compared to earlier datasets.
This perception dataset contains $ \sim $160K training frames, covering diverse conditions including varied weather, lighting, and geographic conditions.
The annotation classes now include vehicles, pedestrians, cyclists, and traffic signs. For comprehensive evaluation, we use both Average Precision (AP) and its heading-aware variant (APH), which incorporates orientation accuracy.
It is noted that: 1) Defining two difficulty levels based on LiDAR point coverage; 2) Employing stricter 3D IoU thresholds (Veh.: 0.7, Ped./Cyc.: 0.5), imposing higher demands on detector localization accuracy. 
Due to computational constraints, our experiments utilize 20\% of Waymo training set, maintaining full evaluation on the complete test set to ensure comprehensive performance validation.

\subsection{Implementation Details}
We adopt the LION-Mamba\cite{liu2024lion} as our baseline, and implement it within the OpenPCDet codebase. 
The main path of WinMamba employs a 2-level feature pyramid architecture ($N=2$), with the channel dimension uniformly set to 64 across all modules and the sampling factor fixed at 2. 
To achieve progressive spatial downsampling, the core window sizes $(W_{x}, W_{y}, W_{z})$ for the four stages are configured as $(13,13,32)$, $(13,13,16)$, $(13,13,8)$, and $(13,13,4)$. 
The auxiliary path strictly adheres to the design outlined in the Method section. 
We retain the positional encoding consistent with the baseline during decoding, treating it as a controlled variable. 
WSF employs a dynamic shift mechanism, where the shift magnitude is given by $\Delta=(W_x/2, W_y/2, W_z/2)$
The network is trained for 80 epochs with a batchsize of 4 on the KITTI Dataset, whilst 36 epochs with a batchsize of 2 on the Waymo Open Dataset. The learning rate is fixed to $3 \times 10^{-3}$.
All experiments were conducted on 2 NVIDIA RTX 3090 GPUs.

\subsection{Main Results}

\textbf{Results on KITTI.} 
Table \ref{table1} presents quantitative comparisons with representative methods on the KITTI benchmark, in order to thoroughly validate the effectiveness of WinMamba.
The experimental results demonstrate that our method achieves superior performance across all difficulty levels for the three target classes: Car, Pedestrian, and Cyclist. 
Compared to the baseline model, the overall mean Average Precision (mAP) increases by 2.7\%, reaching 73.5\%. 
Notably, the detection accuracy for small-scale objects, such as Pedestrians and Cyclists exhibits particularly significant gains. 
This substantial improvement strongly corroborates WinMamba's exceptional capability in modeling fine-grained spatial features within 3D space, highlighting its advantage for object detection in complex scenarios.

\begin{table}[]
\centering
\setlength{\tabcolsep}{3.6mm}
\begin{tabularx}{0.47\textwidth}{cc|ccc|c}
\toprule
\multicolumn{2}{c|}{Modules} & \multicolumn{3}{c|}{Moderate AP} & \multirow{2}{*}{mAP} \\
WSF & AWF & Car & Ped. & Cyc. &  \\ \midrule
 &  & 77.8 & 59.3 & 67.3 & 70.8 \\
\checkmark &  & 77.9 & 62.7 & 68.6 & 72.2 \\
\checkmark & \checkmark & \textbf{78.2} & \textbf{63.3} & \textbf{71.4} & \textbf{73.5} \\ \bottomrule
\end{tabularx}
\caption{WinMamba component analysis on KITTI val. set}
\label{table3}
\end{table}

\textbf{Results on WOD.}
We conduct supplementary experiments on the WOD to verify the generalization capability of WinMamba. The results are presented in Table \ref{table2}.
Due to computational resource constraints, this experiment utilized 20\% of the training set. 
Consequently, the performance evaluation is mainly concentrated on comparisons with the baseline model. 
The results show that WinMamba achieves 73.6\% mAP and 71.5\% mAPH, yielding consistent marginal gains of +0.6\% and +0.5\% compared to the baseline. 
Such advancement substantiates the significant improvement in feature extraction capability afforded by the WinMamba.

\begin{table*}[]
\centering
\setlength{\tabcolsep}{1.9mm}
\begin{tabularx}{\textwidth}{l|c|cc|cc|cc|c}
\toprule
\multicolumn{1}{c|}{} &  & \multicolumn{2}{c|}{Vehicle 3D AP/APH} & \multicolumn{2}{c|}{Pedestrian 3D AP/APH} & \multicolumn{2}{c|}{Cyclist 3D AP/APH} & mAP/mAPH \\
\multicolumn{1}{c|}{\multirow{-2}{*}{Methods}} & \multirow{-2}{*}{Present at} & L1 & L2 & L1 & L2 & L1 & L2 & L2 \\ \midrule
SECOND & Sensors 18 & 72.3/71.7 & 63.9/63.3 & 68.7/58.2 & 60.7/51.3 & 60.6/59.3 & 58.3/57.0 & 61.0/57.2 \\
PointPillars & CVPR 19 & 72.1/71.5 & 63.6/63.1 & 70.6/56.7 & 62.8/50.3 & 64.4/62.3 & 61.9/59.9 & 62.8/57.8 \\
CenterPoint & CVPR 21 & 74.2/73.6 & 66.2/65.7 & 76.6/70.5 & 68.8/63.2 & 72.3/71.1 & 69.7/68.5 & 68.2/65.8 \\
PillarNet & ECCV 22 & 79.1/78.6 & 70.9/70.5 & 80.6/74.0 & 72.8/66.2 & 72.3/71.2 & 69.7/68.7 & 71.0/68.5 \\
AFDetV2 & AAAI 22 & 77.6/77.1 & 69.7/69.2 & 80.2/74.6 & 72.2/67.0 & 73.7/72.7 & 71.0/70.1 & 71.0/68.8 \\
PillarNeXt & CVPR 22 & 78.4/77.9 & 70.3/69.8 & 82.5/77.1 & 74.9/69.8 & 73.2/72.2 & 70.6/69.6 & 71.9/69.7 \\
VoxelNeXt & CVPR 23 & 78.2/77.7 & 69.9/69.4 & 81.5/76.3 & 73.5/68.6 & 76.1/74.9 & 73.3/72.2 & 72.2/70.1 \\
FlatFormer & CVPR 23 & 78.6/78.1 & 69.8/69.4 & 82.9/77.5 & 74.3/69.3 & 76.6/75.6 & 73.9/72.8 & 72.7/70.5 \\
FSDv1 &  NeurIPS 22 & 79.2/78.8 & 70.5/70.1 & 82.6/77.3 & 73.9/69.1 & 77.1/76.0 & 74.4/73.3 & 72.9/70.8 \\
DSVT-Pillar & CVPR 23 & 79.3/78.8 & 70.9/70.5 & 82.8/77.0 & 75.2/69.8 & 76.4/75.4 & 73.6/72.7 & 73.2/71.0 \\
DSVT-Voxel & CVPR 23 & 79.7/79.3 & 71.4/71.0 & 83.7/78.9 & 76.1/71.5 & 77.5/76.5 & 74.6/73.7 & 74.0/72.1 \\
HEDNet & NeurIPS 23 & 81.1/80.6 & 73.2/72.7 & 84.4/80.0 & 76.8/72.6 & 78.7/77.7 & 75.8/74.9 & 75.3/73.4 \\
VoxelMamba &  NeurIPS 24 & 80.8/80.3 & 72.6/72.2 & 85.0/80.8 & 77.7/73.6 & 78.6/77.6 & 75.7/74.8 & 75.3/73.5 \\
FSDv2 & TPAMI 25 & 79.8/79.3 & 71.4/71.0 & 84.8/79.7 & 77.4/72.5 & 80.7/79.6 & 77.9/76.8 & 75.6/73.5 \\
\rowcolor{gray!20} LION-Mamba-L & NeurIPS 24 & 80.3/79.9 & 72.0/71.6 & 85.8/81.4 & 78.5/74.3 & 80.1/79.0 & 77.2/76.2 & 75.9/74.0 \\ 
UniMamba & CVPR 25 & 80.6/80.1 & 72.3/71.8 & 86.0/81.3 & 78.7/74.1 & 80.3/79.3 & 77.5/76.5 & 76.1/74.1 \\ \midrule

LION-Mamba* & NeurIPS 24 & 77.1/76.7 & 68.6/68.2 & 84.0/79.1 & 76.7/72.0 & 76.5/75.5 & 73.7/72.7 & 73.0/71.0 \\
WinMamba(Ours)* & -- & 78.0/77.5 & 69.6/69.1 & 84.3/79.2 & 77.0/72.1 & 77.1/76.1 & 74.3/73.2 & 73.6/71.5 \\ \bottomrule
\end{tabularx}
\caption{Comparison of model performance on the Waymo Open Dataset validation set. Due to computational resource limitations, both WinMamba and LION-Mamba are trained on 20\% of the training set (marked with *). Additionally, to ensure fair comparison with other methods (train with 100\% training data), we adopt LION\cite{liu2024lion} as the reference.}
\label{table2}
\end{table*}

\subsection{Ablation Study}
To systematically evaluate the contribution of each component in WinMamba to overall performance, we conduct validation experiments on the KITTI dataset using a progressive component addition strategy, as detailed in Tabel \ref{table3} \& \ref{table4}. 
Results on the Car, Pedestrian, and Cyclist classes are analysed under the Moderate difficulty setting, with mAP representing the overall evaluation metric. The analysis of the experimental results is as follows:

\textbf{Effectiveness of WSF.}
As shown in Table \ref{table3}, the introduction of WSF significantly improves detection accuracy across all categories, with a notable enhancement for small-scale objects such as pedestrians and cyclists. 
This improvement can be attributed to the inherent voxel sparsity of small-scale objects, where rigid window partitioning amplifies feature discretization, thereby reducing their discriminative power.
Smaller objects are more affected by this window fragmentation compared to larger ones like vehicles. 
WSF compensates for the loss of boundary information, effectively mitigating the disruption of spatial continuity and significantly boosting detection precision for small objects.

\textbf{Effectiveness of AWF.}
As shown in Table \ref{table3}, the overall detection accuracy improves after introducing AWF, although there is a slight decline in the detection accuracy for pedestrians. 
This may be attributed to the auxiliary path in AWF, where the window size is generally larger than that of the baseline.
For linear models, due to the need for serialization operations, a larger window leads to more dispersed semantic information within the window. 
This is less problematic for larger objects, which typically fill most of the window and thus retain concentrated semantic features. In contrast, for smaller objects like pedestrians, the serialized representation within a larger window leads to more dispersed features, potentially impairing detection accuracy.
Nevertheless, overall accuracy has improved, thanks to the information compensation provided during the feature extraction process in AWF.

\begin{table}[t]
\centering
\setlength{\tabcolsep}{2.9mm}
\begin{tabularx}{0.47\textwidth}{cccc|ccc|c}
\toprule
\multicolumn{4}{c|}{Part} & \multicolumn{3}{c|}{Moderate AP} & \multirow{2}{*}{mAP} \\
A & B & C & D & Car & Ped. & Cyc. &  \\ \midrule
 &  &  &  & 77.8 & 59.3 & 67.3 & 70.8 \\
 & \checkmark &  &  & 77.8 & 63.0 & 70.4 & 72.7 \\
\checkmark & \checkmark &  &  & \textbf{78.4} & 60.5 & \textbf{72.3} & 72.6 \\
\checkmark & \checkmark & \checkmark &  & 78.2 & \textbf{63.3} & 71.4 & \textbf{73.5} \\
\checkmark & \checkmark & \checkmark & \checkmark & 77.8 & 60.1 & 67.6 & 71.2 \\ \bottomrule
\end{tabularx}
\caption{AWF component analysis on KITTI val. set}
\label{table4}
\end{table}

\textbf{Subcomponent Analysis of AWF.} 
Table \ref{table4} presents specialized ablations on AWF subcomponents: 
1) Feature Bridging(PartB): Replacing the original feature pyramid’s residual connection with the auxiliary path significantly boosts accuracy for small targets, like pedestrians and cyclists. This demonstrates that the auxiliary path provide more concentrated information representation than direct skip connection.As shown in Tabel \ref{table4} line 4, the residual connection of the original feature pyramid (Part D) leads to a decline in overall performance, indicating that the connection introduces redundant information. Therefore, our method adopts an alternative approach instead of addition.
2) Dual-stream Encoding (PartA) \& Collaborative Decoding (PartC): Vehicles (large-scale) and pedestrians (small-scale) exhibit contrasting responses: encoding layers provide positive gains for large objects but suppress small targets, while decoding layers exhibit the opposite tendency. Their synergistic interaction creates complementary effects, collectively driving comprehensive performance optimization.

\section{Conclusion}
In this work, we introduced WinMamba, a novel window-based 3D feature encoder built upon the Mamba architecture, designed to significantly enhance the multi-dimensional spatial modeling capabilities of linear models. To address inherent limitations of rigid window partitioning, we proposed two key innovations: WSF effectively mitigates cross-window context loss through strategic shifting and fusion, while AWF constructs a high-resolution auxiliary path with adaptive window sizing to align semantic units and recover lost fine details. Extensive experimental validation on the KITTI and Waymo benchmarks, supported by comprehensive ablation studies, demonstrates the efficacy of WinMamba and its constituent fusion mechanisms. Our results confirm that WinMamba provides a simple yet powerful approach for advanced 3D feature encoding, establishing a strong new baseline for spatial modeling tasks.

\paragraph{Acknowledgements.} This work was supported in part by the National Natural Science Foundation of China (No.42571514).

\bibliography{main}
\end{document}